\documentclass[journal]{IEEEtran}

\usepackage{enumerate}

\usepackage{amsmath}
\usepackage{amsthm}

\usepackage{graphicx}
\usepackage{subfigure}

\begin{document}

\title{Semi-centralized control for multi-robot formation and theoretical lower bound}

\author{Shuo~Wan, Jiaxun~Lu, Pingyi~Fan,~\IEEEmembership{Senior Member,~IEEE} and Khaled~B.~Letaief{*},~\IEEEmembership{Fellow,~IEEE}\\

\small
Tsinghua National Laboratory for Information Science and Technology(TNList),\\

Department of Electronic Engineering, Tsinghua University, Beijing, P.R. China\\
E-mail: wan-s17@mails.tsinghua.edu.cn, lujx14@mails.tsinghua.edu.cn, ~fpy@tsinghua.edu.cn\\
{*}Department of Electronic Engineering, Hong Kong University of Science and Technology, Hong Kong\\
Email: eekhaled@ece.ust.hk}

\maketitle

\begin{abstract}
Multi-robot formation control enables robots to cooperate as a working group in completing complex tasks, which has been widely used in both civilian and military scenarios. Before moving to reach a given formation, each robot should choose a position from the formation so that the whole system cost is minimized. To solve the problem, we formulate an optimization problem in terms of the total moving distance and give a solution by the Hungarian method. To analyze the deviation of the achieved formation from the ideal one, we obtain the lower bound of formation bias with respect to system's parameters based on notions in information theory. As an extension, we discuss methods of transformation between different formations. Some theoretical results are obtained to give a guidance of the system design.\\

\end{abstract}

\begin{IEEEkeywords}
position arrangement, formation bias, leader-follower, formation conversion
\end{IEEEkeywords}

%
\IEEEpeerreviewmaketitle

\section{Introduction}
%
%
%
%
\IEEEPARstart{R}{ecent} years, multi-robot moving in a formation are replacing complex single robots for both civilian and military applications. The multi-robot system has many advantages. By simply replacing broken robots, the overall system performance will not be degraded largely. In this way, it can reduce the complexity by decomposing complex tasks into small ones \cite{nanjanath2010repeated}. Furthermore, multi-robot system can complete more complex tasks with a lower cost. In military applications, robots can cooperate as a working group to complete surveillance tasks \cite{young2016survey} or do spying work in adversarial areas based on techniques in \cite{mur2015orb}\cite{deutsch2016framework}\cite{7831359}. In civilian applications, automatically driving car system may be helpful in the intelligent transportation systems.

Formation control is a critical problem for robots to coordinate. To complete a certain task, we often need robots to move by maintaining a sequential formations. For example, when cooperatively transporting a large item, certain formations are required, which enable them to complete the task \cite{chaimowicz2001architecture}. In this process, the formation should be reached as fast as possible to improve the working efficiency. Furthermore, the deviation of achieved formation from the exactly required one should be constrained in a reasonable range to ensure the task's completion. In practice, we also need to change formation according to different tasks and the process should be fast as well.

Multi-robot formation usually requires several basic conditions or assumptions. Here, it assumes that: 1) Robots are able to set relative positioning of others around them by using the assistant essential information to guide themselves \cite{krajnik2014practical}\cite{todescato2016multi}; 2) Robots are able to control themselves to reach their desired places; 3) The basic communication capability is necessary, so that robots are able to cooperate with each other \cite{wang2016multi}\cite{koutsoubelias2016coordinated}.

The multi-robot formation controlling approaches are mainly divided into three classes: decentralized control, centralized control and the combination of centralization and decentralization, refer to as semi-centralized control. In centralized control, there is a central agent monitoring the whole environment and controlling all robots' motion to reach a formation. The agent can be either a computer or a robot. In decentralized control system, there is no central agent. Each robot makes their own decision. They monitor the environment by themselves and exchange their observations with others by communication. In the semi-centralized control method, there exists a central agent distributing tasks and sending commands to robots and each robot reaches its goal by decisions of their own.

To cope with different application scenarios, researchers presented several different methods based on different assumptions and requirements. In \cite{kloder2006path}, it discussed the problem of planning collision-free paths for permutation-invariant multi-robot formations. In \cite{yu2016robot}, it presented a formation controller for a scalable team of robots where communications are unavailable and sensor ranges are limited. Robots can reach a locally desired formation and change gradually to a globally desired one. In \cite{sun2013distributed}, it proposed a method of distributed estimation and control for preserving formation rigidity of mobile robot teams, which is a kind of virtual structure approach. In \cite{karimoddini2013hybrid}\cite{karimoddini2011hybrid}, they proposed a leader-follower controller for multi-robot teams. In \cite{7981341}, it developed a suboptimal controller for a leader-follower formation problem of quadrotors with the consideration of external disturbances. In \cite{beard2001coordination}, it combined centralization and decentralization methods and proposed a coordination architecture for spacecraft formation control. In \cite{qian2016hierarchical}, it presented an approach for formation control of autonomous vehicles traversing along a multi-lane road with obstacles and a certain degree of traffic.

However, to the best knowledge of us, the previous works didn't discuss how to arrange each robot with a position in the formation. Instead, this is usually requirement from applications. Furthermore, they also didn't present theoretical analysis of the achieved formation compared to the exactly required one.

To solve the problem of arranging robots in a formation, we transform the problem to a task distribution problem and solve it with the Hungarian method. Then we define the formation bias. To estimate the bias, we apply notions in information theory. We do an estimation of the Bayes risk by calculating the mutual information between measured value and true value to solve the problem.

In this paper, we focus on leader-follower method which is more similar to human's action mode when standing in a formation. Each robot has a reference and follows it until the robot arrives at the right position. Based on this, we first propose a method to obtain an optimal solution for arranging robots to positions in a given formation by minimizing the predicted total moving distance. Then we also consider the achieved formation's bias to the ideal one. Applying methods in decentralized estimation, we obtain a lower bound of the formation bias with respect to system parameters. This can guide us to design the system according to user's requirements. Finally, we discuss the transformation between different formations. We optimize the choice of new formation center and confirm it by simulation.

Our discussed system belongs to the third class. It has a central agent to send commands to robots. Each robot receives the commands and reaches its goal by itself. In the process, they detect possible collisions and avoid them. Each robot can measure the relative position of its reference and follow it to reach the formation.

The rest paper is organized as follows: A brief problem and notation statement are given in Section \uppercase\expandafter{\romannumeral2}. In Section \uppercase\expandafter{\romannumeral3}, we give method to arrange robots to positions in the formations by transforming it into a distribution optimization problem. In Section \uppercase\expandafter{\romannumeral4}, we give a depiction of the leader-follower method in our considered system. A lower bound of the formation bias is derived in Section \uppercase\expandafter{\romannumeral5}. The proof is given in Section \uppercase\expandafter{\romannumeral6}. Afterwards, we discuss formation conversion in Section \uppercase\expandafter{\romannumeral7}. Some simulation results are presented in Section \uppercase\expandafter{\romannumeral8}. Finally, we give the conclusion.

%

\section{Problem statements and relevant notations}
Consider a group of robots, each has a serial number ranging from 1 to $n$. Their initial positions $\{x_{i}(0)|i=1,2, ......, n\}$ are randomly distributed. Their motion can be described as follows:
\begin{equation}
x_{i}(k+1)=x_{i}(k)+v_{i}(k)
\end{equation}
where $x_{i}(k)\in R^{2}$ is the position of robot $i$ at time slot $k$ with $i\in\{1,2,......,n\}$ and $\|v_{i}(k)\|\leq U_{\rm max}$. $U_{\rm max}$ is the maximum of velocities. $v_{i}(k)$ is the speed of robot $i$.
Given a formation, the robots need to move to reach it, during which there should not have any collision.

In the real world, there are many different demands for robot formation configuration. As for the square formation, we tend to find a leading robot and let other robots follow it to reach a formation. Sometimes, we want the center of the formation to be at a certain position. Thus, we may let robots keep up with the center to move. The center can be given previously by the planner or it can be decided by the system automatically.

The formation is represented by $F=\left \{ f_{1},f_{2},......,f_{n} \right \}$, where $f_{i}$ is the position of the $i$-th node in the formation. Considering the formation on 2D plane, we have $f_{i}\in R^{2}$. Besides, we also let origin to be the center of the formation, which means $\sum_{i=1}^{n}f_{i}=0$. The formation $F$ is predefined by users, which is used by the system as reference to produce controlling forces for reaching the formation.

Having those positions in the formation, the robots have to be arranged according to them. The arrangement is denoted by $D=\{ d_{1},d_{2},......,d_{n}\}$, where $d_{i}\in\{ 1,2,......,n\}$. $d_{i}$ represents the serial number of the $i$-th robot's position in the formation.

When configuring a square formation, we assume there exists a leading robot which is chosen by the controlling center. Each robot keeps up with another robot according to the arrangement mentioned above, while some robots will follow the leading robot directly. The leading robot's serial number is denoted as $i_{lead}$. Besides, we define $H=\left \{ h_{1},h_{2},......,h_{n} \right \}$, where $h_{i}\in\left \{ 1,2,......,n \right \}$ represents the serial number of the robot followed by robot $i$. In addition, we also define $P=\left \{ p_{1},p_{2},......,p_{n} \right \}$, where $p_{i} \in R^{2}$ represents the required relative position in the formation between robot $i$ and the robot it's following. $p_{i}$ can obviously be calculated by $p_{i}=f_{d_{h_{i}}}-f_{d_{i}}$.
Assuming that robot $i$ can measure the relative position of robot $h_{i}$ to itself, the robot can move to adjust the relative position. If for all $i\neq i_{lead}$, there is $x_{i}(k)-x_{h_{i}}(k)=p_{i}$, the formation is reached at time slot $k$. Besides, if the robots collide with another, they may break down. So our controlling method should avoid such event happen.

When requiring the formation to be around a center, we may not require to find a leader. Assuming robots can know the relative position between the center and themselves, they can move to adjust the relative position to achieve the formation. Denoting the position of the center to be $C_{x}$, if at time slot $k$, there are $x_{i}(k)-C_{x}=f_{i}$ for all $i\in\left \{ 1,2,......,n \right \}$, the formation is reached at time slot $k$.

The movement of the robots should be synchronized so that equation (1) can be meaningful to describe the robots' motion \cite{zeng2014resilient}. Assuming that the center or the leading robot can act as the synchronizer, we can achieve this. Besides, the speed should be adjusted from $v_{i}(k-1)$ to $v_{i}(k)$ at time slot $k$. We assume this can be achieved by a motion controller on the robot. As mentioned before, the controlling should be based on the ability of measuring the relative position of other robots \cite{panagou2014cooperative}. We assume the on board sensor and communication system will help to achieve this.

\section{Optimization of formation mapping}
As mentioned, the formation can be described by $F=\left \{ f_{1},f_{2},......,f_{n} \right \}$. Besides, the initial positions of robots are randomly deployed as $X(0)=\left \{ x_{1}(0),x_{2}(0),......,x_{n}(0) \right \}$. Given $X(0)$, we can optimise the arrangement of the robots to the position in the formation described by $D=\left \{ d_{1},d_{2},......,d_{n} \right \}$. The aim of the optimization is to minimize the robots' total moving distance.

Assuming there is a center controller, it will collect the robots' initial position $X(0)$ and calculate the arrangement $D$ according to $X(0)$ and formation $F$. Then it will broadcast the arrangement to robots.

\subsection{Cost Function}
As previously stated, searching for the optimal arrangement $D$ requires minimizing the total moving distance. Moreover, if all the robots choose a position ensuring the shortest path, the total consuming time will be smaller and the moving process seems more reasonable. Before the optimization, we should set up a cost function to describe the moving distance.

To get the cost function, one can estimate the distance between the initial position and the final position in the formation. However, in real system, the moving distance is actually very complex to describe, as the robots may get away the calculated path in the planning stage to avoid other robots or obstacles. For example, when it comes to square formation, each robot follows another one. If it is not following the leader, its destination will keep moving. So the actual path may not be straight, which will result in an extra length  compared to the estimated distance. In this way, the real moving distance may be very difficult to be represented. As an alternative, we use the estimated distance to get a near optimal result. We assume that the robots are not very dense, and the collision does not frequently happen. Besides, considering the actual curve path will make the problem much more complex, which is not worth to do so. Therefore using the estimated distance may be a good selection in the initial planning stage.

Choosing the leader, others will move to the right position relative to it. Besides, $f_{d_{i_{lead}}}$ should also be the leading position in the square formation. Given the leading robot's position $x_{i_{lead}}$ and assuming that the formation is reached at time slot $k$, the $ i$-th robot's current position is $$x_{i}(k)=x_{i_{lead}}-f_{d_{i_{lead}}}+f_{d_{i}}$$ Therefore the estimated moving distance of robot $i$ should be $$C_{i}= ||x_{i}(0)-x_{i}(k)||=||(x_{i}(0)-f_{d_{i}})-(x_{i_{lead}}-f_{d_{i_{lead}}})||$$
So we can get the cost function for robot $i$.
\begin{equation}
C_{t}(i)=C_{i}^{2}=||(x_{i}(0)-f_{d_{i}})-(x_{i_{lead}}-f_{d_{i_{lead}}})||^{2}
\end{equation}

When it requires to reach a formation around a center, it is not need to choose a leading robot. Given a center with position $C_{x}$, each robot may move according to $C_{x}$. Assuming that collision does not happen frequently, the extra path length caused by avoiding possible collisions can be ignored first. Besides, each robot can move according to the fixed center, which means that without considering the few collisions, the path is straight with high possibility. In this way, if the formation is reached at time slot $k$, the right position for robot $i$ will be $$x_{i}(k)=C_{x}+f_{d_{i}}$$ Therefore the estimated moving distance of robot $i$ is $$C_{i}=||x_{i}(0)-x_{i}(k)||=||(x_{i}(0)-f_{d_{i}})-C_{x}||$$
The corresponding cost function for robot $i$ is defined as
\begin{equation}
C_{t}(i)=C_{i}^{2}=||(x_{i}(0)-f_{d_{i}})-C_{x}||^{2}
\end{equation}

\subsection{Optimization problem and Hungarian method}
\subsubsection{Problem Statement}
Considering the square formation with a leader, the optimization problem can be formulated as follows by using cost function (2).
\begin{equation}
D=\underset{D}{\rm argmin}\sum_{i=0,i\neq i_{lead}}^{n} ||(x_{i}(0)-f_{d_{i}})-(x_{i_{lead}}-f_{d_{i_{lead}}})||^{2}
\end{equation}

When it comes to reaching a formation around a center, the cost function is updated by equation (3) and the optimization problem is formulated as
\begin{equation}
D=\underset{D}{\rm argmin}\sum_{i=0}^{n} ||(x_{i}(0)-f_{d_{i}})-C_{x}||^{2}
\end{equation}

In fact, with semi-centralized control mode, the formation can be divided into the two scenarios above along the time. In the first phase, all robots are far from its required positions. So selecting a leader may be a good way to move close to the required positions. Then the second phase is triggered, one can select the mode where the center of formation is known. Now, we first introduce the Hungarian Method by Kuhn in \cite{kuhn1955hungarian}\cite{kuhn1956variants}.

\subsubsection{Review of Hungarian Method}
Considering a general assignment problem, there is an $n$ by $n$ cost matrix $A=[a_{ij}]$, where $a_{i,j}$ satisfies $a_{ij}\geq 0$. The aim is to find a set $j_{1}, j_{2},......, j_{n}$, which is a permutation of $1, 2,......, n$, so that the sum $C_{t} = r_{1,j_{1}}+r_{2,j_{2}}+......+r_{n,j_{n}}$ is minimized.

As for matrix $A$, if there are zero elements existing both in different rows and columns, one can get the assignments easily, and the $C_{t}$ mentioned above is obviously zero. However, $A$ doesn't necessarily contain enough such zero elements. So the matrix have to be transformed by the Hungarian Method. Two basic theorems are introduced here, refer to Lemma 1 and 2, respectively.

\newtheorem{lemma}{Lemma}
\begin{lemma}
Koning's theorem \cite{koniggrafok}: In any bipartite graph, the number of edges in a maximum matching equals the number of vertices in a minimum vertex cover.
\end{lemma}
\newtheorem{remark}{Remark}
\begin{remark}
In Lemma 1, bipartite graph refers to those whose vertices can be partitioned into two sets such that each edge has one endpoint in each set. A vertex cover in a graph is a set of vertices that includes at least one endpoint of every edge, and a vertex cover is minimum if no other vertex cover has fewer vertices. A matching is a set of edges none of which share an endpoint. If no other matching has more edges, a matching is maximum \cite{bondy1976graph}. Based on Lemma 1, there is a conclusion that the minimum number of rows or columns needed to contain matrix A's all zero elements is equal to the maximum number of zeros that can be chosen, in which none of two zero elements is on the same line.
\end{remark}
\begin{lemma}
    the distribution problem is unchanged if the matrix A is replaced by $A^{'} = (a_{ij}^{'})$, with $a_{ij}^{'} = a_{ij}-u_{i}-v_{j}$ for constants $u_{i}$ and $v_{j}$, and $i,j=1,2,......n$ \cite{kuhn1956variants}.
\end{lemma}

With Lemma 2, one can transform matrix $A$ to obtain some zero elements. With Lemma 1, one can find the independent zero elements and minimize $C_{t}$. If there are not enough zero elements, one can continue to transform $A$ with Lemma 2 until success. The detailed procedure can refer to \cite{kuhn1955hungarian}\cite{kuhn1956variants}.

\subsubsection{Problem Transformation and Solution}
The formation mapping problem can be transformed to the assignment problem and solved by the Hungarian Method.

As for the optimization problem (4), given robot $i \neq i_{lead}$, its assigned position's serial number is $d_{i}$. Let $d_{i} = j$, we can have the following equation:
\begin{equation}
a_{ij} = ||(x_{i}(0)-f_{d_{i}})-(x_{i_{lead}}-f_{d_{i_{lead}}})||^{2}
\end{equation}
where $a_{ij}$ is an element in the cost matrix $A$. It represents the cost of arranging robot $i$ to the position with serial number $j$. In the considered system, the leading robot's position should be the explicit reference marker in the square formation. Therefore $d_{i_{lead}}$ can be fixed. However, to calculate $a_{ij}$, we should select the leading robot $i_{lead}$. The value of $a_{ij}$ is then relevant to $i_{lead}$, which means we can calculate a cost matrix $A$ given an $i_{lead}$. In this case, $A$ is $(n-1)$ by $(n-1)$, for the leader is excluded. With $A$, we can use the Hungarian Method to get an optimal arrangement conditionally for a given $i_{lead}$. If we let each robot to be the leader and separately calculate the result, an optimal arrangement can be finally acquired  by comparison of the conditional minimum cost results.

Note that in the optimization problem (5), given $d_{i}=j$ for robot $i$, the cost matrix $A$ is calculated by the following equation:
\begin{equation}
a_{ij} = ||(x_{i}(0)-f_{d_{i}})-C_{x}||^{2}
\end{equation}
That is, given $C_{x}$, the cost matrix $A$ can be acquired. Then the optimal arrangement can be directly achieved by the Hungarian Method. The result is correlated with $C_{x}$. In the considered system, $C_{x}$ is usually predetermined.

\section{Moving strategies}
Given $n$ robots' initial position $x(0)$, formation $F$ and calculated formation mapping $D$, the robots can move to reach the formation. The moving strategy should not only achieve the goal, but also prevent possible collisions.

A good approach to solve the problem is to imitate human. When people need to stand in a formation, they tend to find a reference and move to the exactly desired position relative to it. During the process, if two or more people approach the same position, they may slow down or deviate to avoid possible collisions. The reference can be selected as the leader or other people beside them in the formation. When people want to reach a formation around a center, they can also choose the center as the reference. To conclude, this is actually a leader-follower mode. As for robots, they can also follow such mode. Therefore a polar partitioning model \cite{karimoddini2013hybrid}\cite{karimoddini2011hybrid} can be applied and extended to our scenario.

\begin{figure}[!h]
  \centering
  \includegraphics[width=3in]{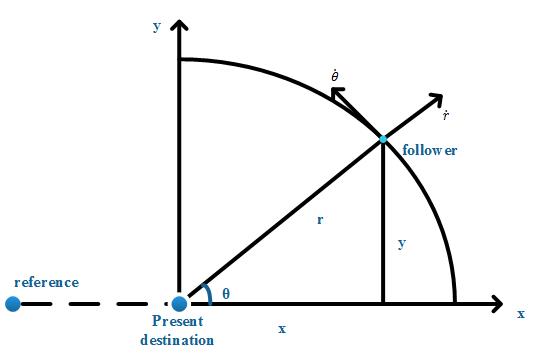}
  \caption{ leader follower model and Circle coordinate system}\label{}
\end{figure}
\begin{figure}[!h]
  \centering
  \includegraphics[width=1.5in]{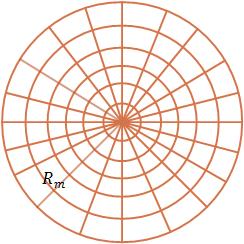}
  \caption{ A partitioned circle whose center is the present destination }\label{}
\end{figure}

  Fig. 1 represents the coordinate system. The follower can measure the leader's position. Together with the relative position in the formation, it can calculate a currently desired destination, which is the origin of the polar coordinates. The follower can measure its destination's position and locate itself in the polar coordinates.

Considering a follower located in circle $R_{m}$ with the radius $R$, the center of $R_{m}$ is the follower's current destination. Fig. 2 shows that $R_{m}$ is divided by curves
\begin{equation}
\left \{ r=r_{h} | r_{h}=\frac{R}{n_{r}-1} (h-1), h=1, 2, ......, n_{r} \right \}
\end{equation}
\begin{equation}
\left \{ \theta=\theta_{j} | \theta_{j}=\frac{2\pi }{n_{\theta}-1} (j-1), j=1, 2, ......, n_{\theta} \right \}
\end{equation}
 A region is denoted by
 \begin{equation}
 R_{hj}=\left \{ x=(r,\theta) | r_{h}\leq r\leq r_{h+1}, \theta(j)\leq \theta\leq \theta(j+1) \right \}
 \end{equation}
 whose boundaries are given in (8) and (9).

Supposing a follower is denoted by robot $i$, at time slot $k$, let $x_{i}(k)\in R_{hj}$. The motion of robot $i$ is described by (1). Then the moving strategies should be
\begin{equation}
x_{i}(k+1) \in R_{(h-1)j}\: \: \: \: (h \neq 1)
\end{equation}
\begin{equation}
v_{i}(k) = 0\: \: \: \: (h = 1)
\end{equation}
 However, if it detects a possible collision in $R_{(h-1)j}$, it will choose to move to $R_{h(j+1)}$ at time slot $(k+1)$. If $R_{h(j+1)}$ is also possible for collision, it will choose $R_{h(j-1)}$. If no possible moving is safe, it will stop for a time slot. When robot $i$ detects possible collision with robot $j$ and $i>j$, only robot $i$ will deviate from the path and robot $j$ will not be affected. In this way, it can improve the system's efficiency intuitively. Therefore the formation can be reached if each robot follows this procedure to find its destination.

 \section{Theoretical analysis of formation bias}
 Based on the procedure described previously, the formation can be finally reached. However, the measurement of position may not be accurate. Besides, the moving strategies need to partition the space, which will cause inaccuracy. So robots' final position may have a bias compared to the exactly required one, which causes a formation bias.

 In this section, we provide a theoretical analysis of the formation's bias. We first introduce the notion of the formation bias and analyze possible issues causing it. Then we give a lower bound of it, which is related to the system's parameters. Finally we use simulations to check the theoretical results.
\subsection{Formation bias definition}
In the system, each robot should measure the position of its destination, including relative distance $r$ and relative angle $\theta$. Then the robot need to quantify $r$ and $\theta$ to complete the space partition. The measurement of $r$ and $\theta$ may bring some bias. Besides, the quantification will also bring bias to the final formation.

Assuming that robot can move from one region to another accurately, the bias mainly results from measurement and quantification. Besides, the angle $\theta$ represents the direction from which the robot approaches its destination. This does not contribute to the final formation bias. So we mainly consider bias caused by $r$ here.

Supposing the formation is reached at time slot $k$, the robots' destinations are denoted by $DS = \left \{ ds_{1}, ds_{2}, ......, ds_{n} \right \}$. For robot $i$, its position bias $Er(i)$ is
\begin{equation}
    Er(i)=|x_{i}(k)-ds_{i}|
\end{equation}
Therefore the formation bias $Er$ can be defined as
\begin{equation}
    Er = \frac{1}{n}\sum_{i=1}^{n} Er(i)
\end{equation}
which represents the mean deviation of robots. The bias is mainly from the measurement of distance $r$. We will analyze the bias of $r$ to give an estimation of $Er$.

\subsection{Problem Modeling}
A model of decentralized estimation with single processor is first considered. In this process, the estimator can't gain direct access to the parameter of interest. It can receive the measurement from a local sensor. The sensor can make several measurements about the parameter. After acquiring the samples, it will quantize them and send them to the estimator for analysis. Finally, it makes an estimation about the parameter.

The estimation performance can be evaluated by the distortion, which is represented by a function of the parameter value and its estimated value. The minimum possible distortion is defined as Bayes risk. In \cite{xu2016information}, it gives the lower bounds of the Bayes risk for the estimator problems.

\begin{figure}[!h]
  \centering
  \includegraphics[width=3.5in]{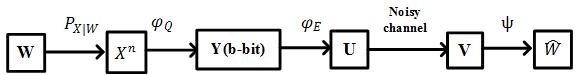}
  \caption{ The model of decentralized estimation with single processor }\label{}
\end{figure}

Fig. 3 shows the decentralized estimation system. $W$ is the parameter of interest, which is measured by local processor. $W$ is derived from a prior distribution $P_{W}$. Given $W=w$, the sensor can get a sample $X$ generated from the distribution $P_{X|W=w}$. Typically, the sensor gets $n$ samples independently according to the distribution $P_{X^{n}|W=w}$, where $X^{n} = \left \{ X_{1}, X_{2}, ......, X_{n} \right \}$. After that, the system uses quantification function $\varphi _{Q}$ to map the observed message $X^{n}$ into a $b$-bit message
\begin{equation}
    Y = \varphi _{Q}(X^n)
\end{equation}
Then the encoder uses the coding function $\varphi _{E}$ to transform $Y$ into codeword
\begin{equation}
    U= \varphi _{E} (Y)
\end{equation}
$U$ is transmitted through a noisy channel and received as $V$.
Finally, the estimator calculates
\begin{equation}
    \hat{W} = \psi (V)
\end{equation}
as the estimates of $W$. Acquiring this estimation model, we can use it to estimate our formation bias.
Before this, we should fit our problem to this model.
\begin{enumerate}[*]
\item $W$ is defined as the robot's distance variable from its present destination and $w$ is its value. We consider robots in the circle $R_{m}$. Assuming $w$ satisfies $w \in [0,l_{0}]$ in practical system, we have $l_{0} < R$.
\item  $X^{n}$ are $n$ independently drawn samples of $W$. The sensor on the robot measures the distance $r$ for $n$ times. The $n$ independent measures are denoted as $\left \{ x_{1}, x_{2}, ......, x_{n} \right \}$.
\item After acquiring the measurement, the system will use curves (8) to divide $R$ and locate the measured distance into one of the regions. This is a process of quantification. The number of divisions $n_{r}$ represents the quantization precision. The result is $b$-bit length $Y$ in the system.
\item In our system, the controller can gain direct access to $Y$. There is not process through the noisy channel. So we don't need to make the channel coding and transmit it to the receiver. Therefore we have $V = U = Y$.
\end{enumerate}

Before the analysis, the relative distributions and equations should also be defined.

$P_{W}$ is a prior distribution of $W$. In this system, $W$ represents the distance $r$. As mentioned above, robot $i$ is in circle $R_{m}$, whose center is the robot's present destination. As mentioned above, we assume $W \in [0,l_{0}]$. $l_{0}$ is an empirical constant of the system. To ensure the quantification being effective, there is $R > l_{0}$. To calculate $P_{W}$, we assume that robot $i$ has an equal possibility to be located at each point in the circle with the radius $l_{0}$. Therefore
\begin{equation}
    P(W < r) = \frac{S_{r}}{S_{l_{0}}} = \frac{\pi r^2}{\pi l_{0}^2} = \frac{r^2}{l_{0}^2}
\end{equation}
where $P(W<r)$ represents the probability that the robot's distance to its destination is smaller than $r$. $S_{r}$ and $S_{l_{0}}$ respectively represents the acreage of the circle with the radius $r$ and $l_{0}$.

Acquiring the distribution function, we need to calculate the density function.
\begin{equation}
    P(W=r) =  \frac{{\rm d}P(W<r)}{{\rm d}r} =  \frac{2r}{l_{0}^{2}}
\end{equation}

$P_{X|W}$ represents the distribution of measurement $X$ conditioned on $W$. Given $W=w$, the measurement random variable $X$ is generated according to the distribution $P_{X|W=w}$. In this system we choose Gaussian distribution as $P_{X|W=w}$.
\begin{equation}
    P_{X|W=w}=\frac{1}{\sqrt{2\pi \sigma ^{2}}}\times e^{\frac{(X-w)^{2}}{-2\sigma ^{2}}}
\end{equation}
where the real value of the parameter is $w$, which is also the mean value of the Gaussian distribution. $\sigma$ is the measurement variance.

In the considered system, it assumes that the sensor measures the distance for $n$ times independently. The mean and variance value are with the same $w$ and $\sigma$. In this case, a joint Gaussian distribution is applied here to represent $P_{X^{n}|W=w}$.
\begin{equation}
    P_{X^{n}|W=w}=\frac{1}{(\sqrt{2\pi \sigma ^{2}})^{n}}\times e^{\frac{\sum_{1}^{n}(x_{i}-w)^{2}}{-2\sigma ^{2}}}
\end{equation}

The measured distance is not directly used by the controller. It has to be quantified first. The distance $r$ satisfies $r \in [0,l_{0}]$ and there is $l_{0} < R$, then $[0,R]$ is the quantitative range. The range is divided into $n_{r}$ regions, so the quantization bit $b$ satisfies
\begin{equation}
    b = {\rm log}_{2}(n_{r})
\end{equation}

In this way, the distributed estimation system is applied to the formation bias estimation problem. Then we can use theories in distributed estimation to give a lower bound on the Bayes risk of the estimation. This will guide us to choose the system's parameter.

\subsection{Lower Bounds Estimation}
As shown in Fig. 3, the parameter of interest is $W$, which is derived from a prior distribution $P_{W}$. After sampling and quantification, the parameter is estimated as $\hat{W} = \psi (V)$. By using a non-negative distortion function $l : W \times W \rightarrow R^{+} $, we can define the Bayes risk of estimating $\hat{W}$ as
\begin{equation}
    R_{B}=\underset{\psi}{\rm inf }\:\: {\rm E}(l(W,\psi (V)))
\end{equation}
where the distortion function $l$ is of the form $l(w,\hat{w}) = ||w-\hat{w}||^{q}$ and $q \geq 1$. In this paper, $W$ represents the distance $r$. Therefore function $l$ is defined on $R^{1}$. In \cite{xu2016information}, it provides lower bounds on the Bayes risk for estimation. Then based on the above model and distributions, we can obtain a lower bound of $R_{B}$ and derive the formation's bias.

\newtheorem{mypro}{Theorem}
\begin{mypro}
In a multi-robot formation control system, we measure the distance $n$ times with a measuring variance $\sigma$. The radius of controlling circle $R_{m}$ is $R$ and the quantification rate is set as $b$. If the measured distance satisfies $r \in [0,l_{0}]$, we can derive the lower bound of $R_{B}$:
\begin{equation}
    R_{B} \geq \frac{l_{0}}{2e} {\rm max}\left \{ 2^{-( {\rm log}\frac{nl_{0}}{2\sigma } +\frac{1}{2} -\frac{1}{2}{\rm log}\left ( 2\pi e \right ))}, 2^{-\left ( 1-e^{-\frac{nl_{0}^{2}}{2\sigma ^{2}}} \right )b} \right \}
\end{equation}
where $l(w,\hat{w}) = ||w-\hat{w}||$.
\end{mypro}

\begin{remark}
The above $R_{B}$ describes measuring bias of the distance. This bias describes the robot's deviation from its right position in the formation. The whole formation bias is the mean value of each robot's deviation as defined in (14). Therefore the lower bound of $R_{B}$ can be viewed as the lower bound of formation bias.
\end{remark}

\section{Proof of lower bound}
In this section, we give proof of the lower bound in Theorem 1. In \cite{xu2016information}, the author gives theorems to estimate lower bound of Bayes risk. We introduce the theorems as lemmas to analyze the formation bias.

\begin{lemma}
For any arbitrary norm $\left \| \cdot  \right \|$ and any $q \geq 1$, when the parameter of interest $W \in R^{d}$ and $W$ is distributed in $[0,1]$. The Bayes for estimating the parameter based on the sample $X$ with respect to the distortion function $l(w,\hat{w}) = \left \| w-\hat{w}  \right \|^{r}$ satisfies
\begin{equation}
    R_{B}\geq \underset{P_{U|W,X}}{\rm sup}\, \frac{d}{qe}\left ( V_{d}\Gamma (1+\frac{d}{q}) \right )^{-\frac{q}{d}}\, 2^{-\left ( I(W;X|U)-h(W|U) \right )\frac{q}{d}}
\end{equation}
To make problems simple, when we are estimating a real-valued $W$ with respect to $l(w,\hat{w}) = \left \| w-\hat{w}  \right \|$,
\begin{equation}
    R_{B}\geq \underset{P_{U|W,X}}{\rm sup}\,\frac{1}{2e} \, 2^{-\left ( I(W;X|U)-h(W|U) \right )}
\end{equation}
Considering the unconditional version, there's a simpler form
\begin{equation}
    R_{B}\geq \frac{1}{2e} \, 2^{-I(W;X)}
\end{equation}
\end{lemma}
In fact, (27) is an unconditional version of lemma 3, which can give us a lower bound of $R_{B}$. This lower bound is for the case $W \in [0,1]$. If the distribution range is changed, we can multiply a parameter to correct it.

In the single processor estimating system displayed by Fig. 3, the estimator can't gain direct access to the measurement $X^{n}$. $X^{n}$ is quantified and transmitted to the estimator. In fact, the estimation is based on $V$. So we can use unconditional version of lemma 3 to obtain a lower bound of Bayes risk $R_{B}$ by replacing $I(W;X)$ with $I(W;V)$. Now, we need to calculate an upper bound of $I(W;V)$.

\begin{lemma}
In decentralized estimation with a single processor, for any choice of $\phi_{Q}$ and $\phi_{E}$, there is
\begin{equation}
\begin{split}
    I(W,V)\leq {\rm min}\{I(W,X^{n})\eta _{T}, \eta (P_{X^{n}},P_{W|X^{n}})(H(X^{n})\wedge b)\eta _{T},\\ \eta (P_{X^{n}},P_{W|X^{n}})CT \}
\end{split}
\end{equation}
where C is the Shannon Capacity of the noisy channel $P_{V|U}$, and
\begin{equation}
    \eta _{T}=\left \{\begin{aligned}1-(1-\eta \left ( P_{V|U}) \right )^{T}   &   &{ with\,\,feedback}\\ \eta \left ( P_{V|U}^{\bigotimes T} \right )  &  &{ without\,\,feedback} \end{aligned} \right.
\end{equation}
\end{lemma}

In (29), $T$ is the time spent transmitting a message $Y$ through the channel $P_{V|U}$, and $\eta$ is a constant.

On the calculation of $\eta$, it can be described as follows. Given a channel $K$ whose input alphabet is $X$ and output alphabet is $Y$. There is a reference input distribution $\mu $ on $X$. If for a constant $c \in [0,1)$ and any other input distribution $\nu$ on $X$, there is $D(\nu K || \mu K) \leq cD(\nu || \mu)$, we say $K$ satisfies an SDPI at $\mu$.

The SDPI constant of $K$ at input distribution $\mu$ is defined as
\begin{equation}
    \eta \left ( \mu ,K \right ) = \underset{\nu:\nu \neq \mu }{\rm sup}\frac{D\left ( \nu K|| \mu K\right )}{D\left ( \nu ||\mu  \right )}
\end{equation}
The SDPI constant of $K$ is defined as
\begin{equation}
    \eta \left ( K \right )= \underset{\mu }{\rm sup}\, \eta \left ( \mu ,K \right )
\end{equation}

With Lemma 3, one can obtain a lower bound of Bayes risk $R_{B}$ relative to the mutual information $I(W,V)$. With Lemma 4, one can calculate an upper bound of $I(W,V)$ in our formation bias estimating system based on the model in section \uppercase\expandafter{\romannumeral5}.B, and then derive the result of Lemma 3.

In our considered formation bias estimation model, the quantified result $Y$ is directly used for estimation. So the transmitting channel can be viewed as lossless. From the definition of $\eta$ (30) (31), we can obtain
\begin{equation}
    \eta ( P_{V|U}) = 1
\end{equation}
Then according to the definition (29), there is
\begin{equation}
    \eta _{T} = 1
\end{equation}
Besides, our model considers no channel loss. That is, the channel capacity $C$ is infinite. In (28), the upper bound of $I(W,V)$ is the minimum value of three parts. The last part $\eta (P_{X^{n}},P_{W|X^{n}})CT$ is related with $C$. Therefore it can be ignored. Then the upper bound (28) can be simplified to the following
\begin{equation}
    I(W,V)\leq {\rm min}\{I(W,X^{n}), \eta (P_{x^{n}},P_{W|X^{n}})(H(X^{n})\wedge b) \}
\end{equation}
In this considered system, $X^{n}$ is the measurement of the distance between the robot and its present destination. It contains $n$ samples taken from the sensor. Its element $X_{i}$ satisfies $X_{i} \in [0,l_{0}]$. This is a continuous variable. When quantifying it with $b$ bit, it will certainly result in some bias. So $H(X^{n})$ is larger than $b$. That is,
\begin{equation}
    H(X^{n})\wedge b = b
\end{equation}
Therefore the upper bound (34) can be simplified further as
\begin{equation}
    I(W,V)\leq {\rm min}\{I(W,X^{n}), \eta (P_{X^{n}},P_{W|X^{n}})b \}
\end{equation}
Next we should separately estimate $I(W,X^{n})$ and $\eta (P_{x^{n}},P_{W|X^{n}})b$ to finally obtain the upper bound.

Clarke \cite{clarke1994jeffreys} shows that
\begin{equation}
\begin{split}
    I\left ( W,X^{n} \right )= \frac{d}{2}log\left ( \frac{n}{2\pi e} \right ) + h\left ( W \right ) + \frac{1}{2}{\rm E}\left [ log\, det\, J_{X|W}\left ( W \right ) \right ]\\ + o\left ( 1 \right )
\end{split}
\end{equation}
where $h(W)$ is the differential entropy of $W$, and $J_{X|W}\left ( W \right )$ is the Fisher information matrix about $w$ contained in $X$.

From (19), we get the density function of $W$
\begin{equation}
    p(W) = \frac{2W}{l_{0}^{2}}
\end{equation}

The differential entropy of $W$ is
\begin{equation}
\begin{aligned}
    h(W) &= {\rm E}(-log(p(W))) \\
         &= \int_{0}^{l_{0}} -\frac{2W}{l_{0}^{2}}\, log(\frac{2W}{l_{0}^{2}})\, {\rm d}W \\
         &= \frac{1}{2}-log\frac{2}{l_{0}}
\end{aligned}
\end{equation}

Now, we need to calculate the fisher information.
The fisher information can be written as
\begin{equation}
    {\rm det}\, J_{X|W}\left ( W \right ) = -{\rm E}[\frac{\partial ^{2}}{\partial W^{2}}\, log\left ( P_{X|W} \right )]
\end{equation}
From (21), $P(X|W)$ is a joint Gaussian distribution of $n$ independent samples. Then we have
\begin{equation}
    {\rm log}\left ( P_{X|W} \right ) = -\frac{n}{2}{\rm log}\left ( 2\pi \sigma ^{2} \right )-\frac{1}{2\sigma ^{2}}\sum_{i=1}^{n}\left ( x_{i}-W \right )^{2}
\end{equation}
Then
\begin{equation}
\begin{aligned}
    {\rm det}\, J_{X|W}\left ( W \right ) &= -{\rm E}[\frac{\partial ^{2}}{\partial W^{2}}(-\frac{1}{2\sigma ^{2}}\sum_{i=1}^{n}\left ( x_{i}-W \right )^{2})] \\
                                    &= \frac{1}{2\sigma ^{2}}{\rm E}[\frac{\partial }{\partial W}(-2\sum_{i=1}^{n}\left ( x_{i}-W \right ))] \\
                                    &= \frac{1}{2\sigma ^{2}}{\rm E}[2n] \\
                                    &= \frac{n}{\sigma ^{2}}
\end{aligned}
\end{equation}

In this system, since $W \in R^{1}$, we have $d = 1$. By using (39) and (42), one can obtain an estimation of $I(W,X^{n})$.
\begin{equation}
\begin{aligned}
    I\left ( W,X^{n} \right ) &= \frac{1}{2}{\rm log}\left ( \frac{n}{2\pi e} \right )+\frac{1}{2}-{\rm log}\left ( \frac{2}{R} \right )+\frac{1}{2}{\rm log}\left ( \frac{n}{\sigma ^{2}} \right ) + o(1)\\
                              &= {\rm log}\, \frac{nR}{2\sigma }+\frac{1}{2}-{\rm log}\, 2\pi e + o(1)
\end{aligned}
\end{equation}
The first part $I\left ( W,X^{n} \right )$ is estimated. We shall estimate the second part of (36). The critical part of this is to estimate $\eta (P_{X^{n}},P_{W|X^{n}})$. Here, a lemma in \cite{xu2016information} can help us achieve it.

\begin{lemma}
    For a joint distribution $P_{W,X}$, suppose there is a constant $\alpha \in (0,1]$ such that the forward channel $P_{X|W}$ satisfies
    \begin{equation}
        \frac{{\rm d}P_{X|W=w}}{{\rm d}P_{X|W=w^{'}}}\left ( x \right )\geq \alpha
    \end{equation}
    for all $x \in X$ and $w,w^{'} \in W$. \\
    Then the SDPI constants of the forward channel $P_{X|W}$ and the backward channel $P_{W|X}$ satisfy
    \begin{equation}
        \eta \left ( P_{X|W} \right )\leq 1-\alpha
    \end{equation}
    and
    \begin{equation}
        \eta \left ( P_{W|X} \right )\leq 1-\alpha
    \end{equation}
\end{lemma}

In the considered system, we measure the distance independently for $n$ times and get $X^{n}$. So we replace $\eta \left ( P_{W|X} \right )$ with $\eta \left ( P_{W|X^{n}} \right )$.
From (31), we have
\begin{equation}
    \eta \left ( P_{W|X^{n}} \right ) \geq \eta (P_{X^{n}},P_{W|X^{n}})
\end{equation}
Then if (46) holds, we have
\begin{equation}
\begin{aligned}
    \eta (P_{X^{n}},P_{W|X^{n}}) &\leq \eta \left ( P_{W|X^{n}} \right )\\
                                 &\leq 1-\alpha
\end{aligned}
\end{equation}
Therefore the critical step of this part is to estimate $\alpha$.

From (21), $P_{X|W=w}$ is a joint Gaussian distribution, with $w$ as its mean. Then we can use the density function (21) to calculate
\begin{equation}
\begin{aligned}
    \frac{{\rm d}P_{X|W=w}}{{\rm d}P_{X|W=w^{'}}}\left ( x \right ) &= \frac{\frac{1}{(\sqrt{2\pi \sigma ^{2}})^{n}}e^{\sum_{i=1}^{n} -\frac{\left ( x_{i}-w \right )^{2}}{2\sigma ^{2}}}}{\frac{1}{(\sqrt{2\pi \sigma ^{2}})^{n}}e^{\sum_{i=1}^{n} -\frac{\left ( x_{i}-w^{'} \right )^{2}}{2\sigma ^{2}}}} \\
                                                        &= e^{\sum_{i=1}^{n}\frac{\left ( x_{i}-w^{'} \right )^{2}}{2\sigma ^{2}} - \sum_{i=1}^{n}\frac{\left ( x_{i}-w \right )^{2}}{2\sigma ^{2}}}  \\
                                                        &= e^{\frac{\sum_{i=1}^{n}\left ( w-w^{'} \right )\left ( 2x_{i}-w-w^{'} \right )}{2\sigma ^{2}}}
\end{aligned}
\end{equation}

 As each measurement $x_{i}$ is obtained independently. Therefore finding the minimum value of $\left ( w-w^{'} \right )\left ( 2x_{i}-w-w^{'} \right )$ is equivalent to the calculation of minimum value of (49).

\begin{equation}
    h = {\rm min}\left \{ \left ( w-w^{'} \right )\left ( 2x-w-w^{'} \right ) \right \}
\end{equation}
where $x \in [0,l_{0}]$ and $w,w^{'} \in [0,l_{0}]$.

It is easy to see that
\begin{equation}
    \left ( w-w^{'} \right )\left ( 2x-w-w^{'} \right ) = 2(w-w^{'})x-(w^{2}-w^{'2})
\end{equation}
When $w > w^{'}$, (51) increases as $x$ increases. So the minimum value is obtained as $x = 0$. Then we have
\begin{equation}
    h = -(w^{2}-w^{'2})\; \; \; \; \; \; (w > w^{'})
\end{equation}
For $w,w^{'} \in [0,l_{0}]$, we have
\begin{equation}
    h = -l_{0}^{2}\; \; \; \; \; \;(w=0, w^{'}=l_{0})
\end{equation}

Similarly, when $w < w^{'}$, (51) decreases as $x$ increases. So the minimum value is obtained as $x = l_{0}$. Then we have
\begin{equation}
    h = 2(w-w^{'})l_{0}-(w^{2}-w^{'2})
\end{equation}
To calculate the minimum value, we transform (54) as the following.
\begin{equation}
    h=w^{'2}-2w^{'}l_{0}-w^{2}+2wl_{0}
\end{equation}
(55) can be viewed as a quadratic function of $w^{'}$. The minimum value can be reached as $w^{'} = l_{0}$, for $w < w^{'}$. That is, we have
\begin{equation}
    h = -l_{0}^{2}-w^{2}+2wl_{0}\; \; \; \; \; \;(w^{'} = l_{0})
\end{equation}
(56) can be further viewed as a quadratic function of $w$. The minimum value can be reached as $w = 0$. Therefore
\begin{equation}
    h = -l_{0}^{2}\; \; \; \; \; \;(w = 0)
\end{equation}
Together with (53) and (57), we have
\begin{equation}
    {\rm min}\left \{ \left ( w-w^{'} \right )\left ( 2x-w-w^{'} \right ) \right \} = -l_{0}^{2}
\end{equation}
Then there is
\begin{equation}
    \frac{\sum_{i=1}^{n}\left ( w-w^{'} \right )\left ( 2x_{i}-w-w^{'} \right )}{2\sigma ^{2}} \geq -\frac{nl_{0}^{2}}{2\sigma ^{2}}
\end{equation}
Together with (49), we have
\begin{equation}
    \frac{{\rm d}P_{X|W=w}}{{\rm d}P_{X|W=w^{'}}}\left ( x \right ) \geq e^{-\frac{nl_{0}^{2}}{2\sigma ^{2}}}
\end{equation}
That is, (see (44))
\begin{equation}
    \alpha = e^{-\frac{nl_{0}^{2}}{2\sigma ^{2}}}
\end{equation}
Finally, from Lemma 5 and (48), we have
\begin{equation}
    \eta (P_{X^{n}},P_{W|X^{n}}) \leq 1-e^{-\frac{nl_{0}^{2}}{2\sigma ^{2}}}
\end{equation}
Now we have finished estimating the second part of (36). We can get a conclusion
\begin{equation}
\begin{split}
    I\left ( W,V \right )\leq  {\rm min}\left \{ {\rm log}\frac{nl_{0}}{2\sigma } +\frac{1}{2} -\frac{1}{2}{\rm log}\left ( 2\pi e \right ) ,\left ( 1-e^{-\frac{nl_{0}^{2}}{2\sigma ^{2}}} \right )b \right \}
\end{split}
\end{equation}
In (27), we replace $I(W,X)$ with $I(W,V)$. The lower bound in (27) should be corrected by multiplying $l_{0}$. According to the upper bound given in (63), the final lower bound of the Bayes risk can be obtained
\begin{equation}
    R_{B} \geq \frac{l_{0}}{2e} {\rm max}\left \{ 2^{-( {\rm log}\frac{nl_{0}}{2\sigma } +\frac{1}{2} -\frac{1}{2}{\rm log}\left ( 2\pi e \right ))}, 2^{-\left ( 1-e^{-\frac{nl_{0}^{2}}{2\sigma ^{2}}} \right )b} \right \}
\end{equation}

Therefore the proof of theorem 1 is completed.

\section{Formation conversion control}
In the practical use, robots should not only form a formation, but also change to another one to accomplish a different task. The formations are usually planned according to the routine requirements. This includes several different cases.
\begin{enumerate}[*]
\item Robots reach a formation with a leader chosen by the user.
\item Robots reach a formation with a leader at the leading place. The leader is chosen by the system. The leading place is determined by the formation.
\item Robots reach a formation around a center chosen by the planner according to its requirement.
\item Robots reach a formation around a center chosen by the system to optimize the process.
\end{enumerate}

In the following discussion, we assume the new formation's area should be the same as the former one.

In this part, we mainly discuss the transformation between square formation, triangle formation and circle formation. The method can also be applied to other transformations. Besides, we also give a moving strategy for a given center by the planner and a method to adaptively choose a center. We also confirm that the chosen center can minimize the total moving distance.

\subsection{Triangle Formation}
\begin{figure}[!h]
  \centering
  \includegraphics[width=2in]{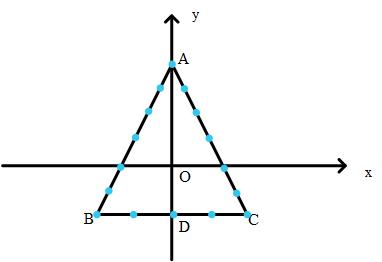}
  \caption{ a triangle formation with robots stand on the black points along the edges  }\label{}
\end{figure}
Fig. 4 shows the structure of a triangle formation with its center on the origin. This is a isosceles triangle, with equal numbers of robots on its two waist edges. From geometry, we know $AO = 2OD$. Excluding robots on the three vertices, we assume there are $x$ robots on one waist edge and $y$ robots on the bottom edge. The total number of robots is $n$, then we have
\begin{equation}
    3+2x+y = n
\end{equation}
Therefore given $y$, we can have the arrangement of robots.

Given the area of the formation as $a \times b$, we can set $$AD = a\,\,\,\,BC = 2b$$ or $$AD = b\,\,\,\,BC = 2a$$ Then the position of $A$,$B$ and $C$ can be determined to locate every point in the formation.

\subsection{Circle Formation}
\begin{figure}[!h]
  \centering
  \includegraphics[width=2in]{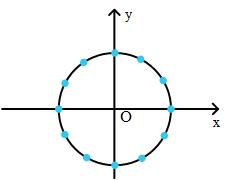}
  \caption{ a circle formation with robots stand on the arc evenly  }\label{}
\end{figure}

Fig. 5 shows a circle formation with its center at the origin. Given the formation's area $S$, we can calculate the radius by $r = \sqrt{\frac{S}{\pi }}$. Then the positions of all points can be determined by partitioning the whole circle equally into $n$ sections.

\subsection{Transformation Strategies}
Given a new formation, the robots need to move from the present one to it.

In Section \uppercase\expandafter{\romannumeral3} and Section \uppercase\expandafter{\romannumeral4}, we have discussed methods to arrange robots in a given formation for arbitrary initial positions for robots and move to reach the formation.
\subsubsection{Center given by planner}
If the center is given and is not far, the robots can move to the proper positions around the center. But if the center is too far, this may be very costly. We should first reduce the distance between the desired center and those robots.

Assuming robots' position at time slot $k$ is $$X(k) = \left \{ x_{1}(k), x_{2}(k), ......, x_{n}(k) \right \}$$
Their present center is
\begin{equation}
    C_{xp} = \frac{1}{n}\sum_{i=1}^{n}x_{i}(k)
\end{equation}
while the desired center given by planner is denoted as $C_{x}$. Then we let all robots move towards the direction of $C_{x}-C_{xp}$ at a given speed. Given a threshold $d_{0}$, if there is
\begin{equation}
    ||C_{x}-C_{xp}|| \leq d_{0}
\end{equation}
the robots are close enough to the center. Then They follow method in section \uppercase\expandafter{\romannumeral3} and section \uppercase\expandafter{\romannumeral4} to reach the formation around the center.
\subsubsection{Center determined by the system}
If the planner don't give the position of the center, the system need first find a center. From (3), the expected moving cost for reaching the formation is
\begin{equation}
    C = \sum_{i=1}^{n} ||(x_{i}(0)-f_{d_{i}})-C_{x}||^{2}
\end{equation}
We need to choose the center to minimize $C$.
\begin{mypro}
For n robots with the initial position $X(0) = \left \{ x_{1}(0), x_{2}(0), ......, x_{n}(0) \right \}$ and a formation $F = \left \{ f_{1},f_{2},......,f_{n} \right \}$ which satisfies $\sum_{i=1}^{n}f_{i}=0$, when choosing the center as
\begin{equation}
    C_{x} = \frac{1}{n} \sum_{i=1}^{n} x_{i}(0)
\end{equation}
the moving cost $C$ is minimized
\end{mypro}

Proof: Defining a function $F$ of $C_{x}$ as
\begin{equation}
    F(C_{x}) = C = \sum_{i=1}^{n} ||(x_{i}(0)-f_{d_{i}})-C_{x}||^{2}
\end{equation}
then
\begin{equation}
\begin{aligned}
    \frac{{\rm d}F}{{\rm d}C_{x}} &= \sum_{i=1}^{n}-2(x_{i}(0)-f_{d_{i}}-C_{x})\\
                     &= -2[\sum_{i=1}^{n} x_{i}(0)-\sum_{i=1}^{n}f_{d_{i}}-nC_{x}]
\end{aligned}
\end{equation}
For $\sum_{i=1}^{n}f_{i}=0$, we have
\begin{equation}
    \frac{{\rm d}F}{{\rm d}C_{x}} = -2[\sum_{i=1}^{n} x_{i}(0)-nC_{x}]
\end{equation}
Let $\frac{{\rm d}F}{{\rm d}C_{x}} = 0$, we have
\begin{equation}
    C_{x} = \frac{1}{n} \sum_{i=1}^{n} x_{i}(0)
\end{equation}
and it's obviously that this is the minimum point.

According to Theorem 2, one can choose the optimal center and follow the methods in Section \uppercase\expandafter{\romannumeral3} and Section \uppercase\expandafter{\romannumeral4} to reach the formation.

\section{Simulations}
In this section, we shall present some simulations for the multi-robot formation control. At the beginning, we present a group of robots reaching a square formation with a leader. Afterwards, we provide a display of robots in square formation changing to circle formation, then to triangle formation, and finally to a circle formation with a far away center. Then the moving cost of our position arrangement strategy in Section \uppercase\expandafter{\romannumeral3} is compared with other position arrangement strategies by simulations. At last, we discuss the relations between formation bias and system parameter. All simulations are run on MATLAB.

\subsection{Formation Illustration}
In this simulation, all robots are in a $1100 \times 1100$ square area. The number of robots $n$ is set to be 15, and their initial positions are generated randomly in a $300 \times 300$ area.

We set the radius of $R_{m}$ as $R=300$, the quantification rate as $b=7$, the area of the formation as $S=28800$ and the variance of measuring the distance as $\sigma=1$. The estimated total moving cost is calculated by definitions in (2) and (3), while the practical moving cost is calculated with simulated data.
\begin{figure}[!h]
\centering
\subfigure[] { \label{Fig:pm108}
\includegraphics[width=0.45\columnwidth]{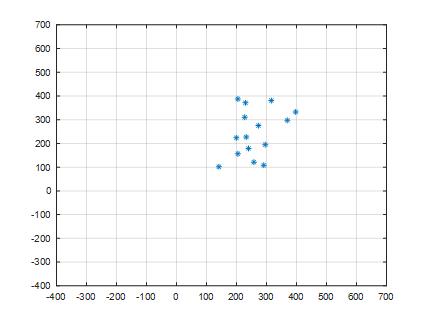}}
\subfigure[]{ \label{Fig:pm1077}
\includegraphics[width=0.45\columnwidth]{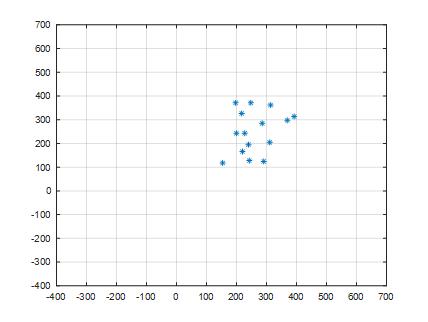}}\\
\subfigure[] { \label{Fig:pm1073}
\includegraphics[width=0.45\columnwidth]{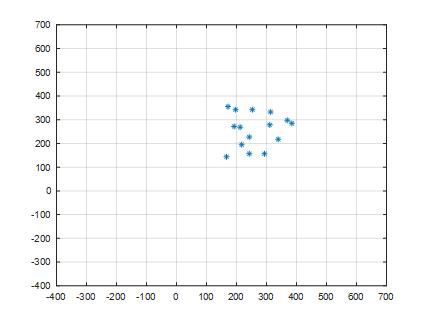}}
\subfigure[]{ \label{Fig:PmaxPosition}
\includegraphics[width=0.45\columnwidth]{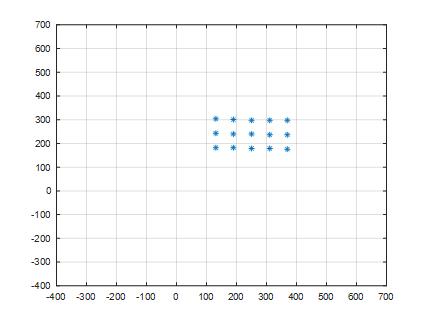}}
\caption{Sub-figures (a) shows a group of robots in random initial position. (b)-(c) show the robots begin to move according to the calculated mapping arrangement. (d) shows robots having reached a square formation.} \label{Fig:PmaxEffects}
\end{figure}

\begin{figure}[!h]
\centering
\subfigure[] { \label{Fig:pm108}
\includegraphics[width=0.45\columnwidth]{fig9.jpg}}
\subfigure[]{ \label{Fig:pm1077}
\includegraphics[width=0.45\columnwidth]{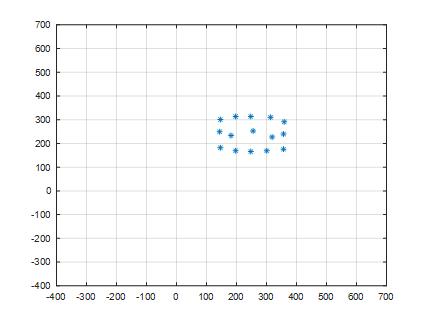}}\\
\subfigure[] { \label{Fig:pm1073}
\includegraphics[width=0.45\columnwidth]{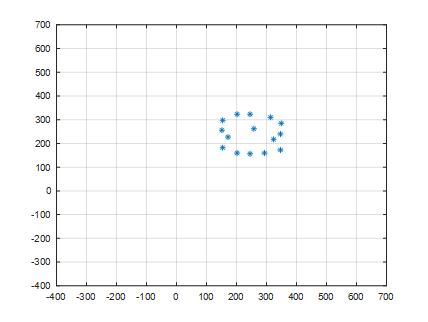}}
\subfigure[]{ \label{Fig:PmaxPosition}
\includegraphics[width=0.45\columnwidth]{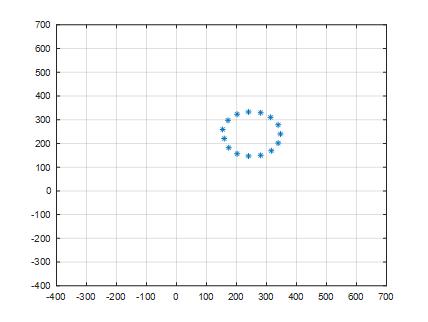}}
\caption{Sub-figures (a) shows a group of robots in a square formation. (b)-(c) show the robots begin to move according to new mapping in circle formation. (d) shows robots having reached a circle formation.} \label{Fig:PmaxEffects}
\end{figure}

\begin{figure}[!h]
\centering
\subfigure[] { \label{Fig:pm108}
\includegraphics[width=0.45\columnwidth]{fig12.jpg}}
\subfigure[]{ \label{Fig:pm1077}
\includegraphics[width=0.45\columnwidth]{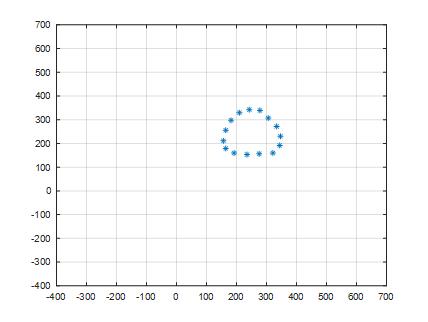}}\\
\subfigure[] { \label{Fig:pm1073}
\includegraphics[width=0.45\columnwidth]{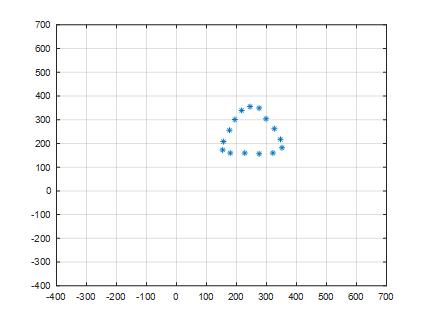}}
\subfigure[]{ \label{Fig:PmaxPosition}
\includegraphics[width=0.45\columnwidth]{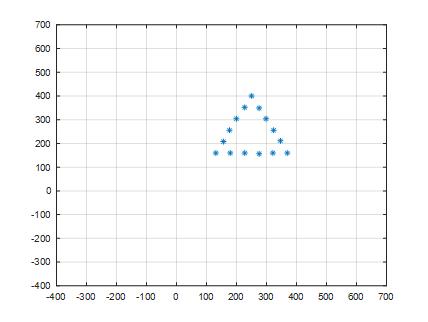}}
\caption{Sub-figures (a) shows a group of robots in a circle formation. (b)-(c) show the robots begin to move according to new mapping in triangle formation. (d) shows robots having reached a triangle formation.} \label{Fig:PmaxEffects}
\end{figure}

\begin{figure}[!h]
\centering
\subfigure[] { \label{Fig:pm108}
\includegraphics[width=0.45\columnwidth]{fig15.jpg}}
\subfigure[]{ \label{Fig:pm1077}
\includegraphics[width=0.45\columnwidth]{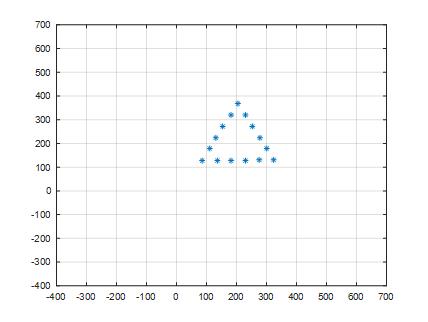}}\\
\subfigure[] { \label{Fig:pm1073}
\includegraphics[width=0.45\columnwidth]{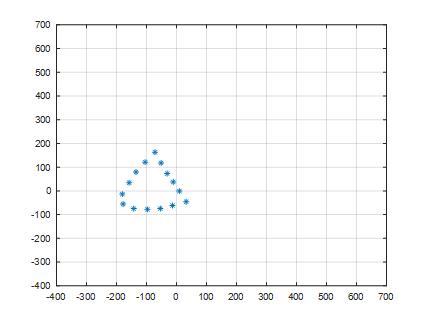}}
\subfigure[]{ \label{Fig:PmaxPosition}
\includegraphics[width=0.45\columnwidth]{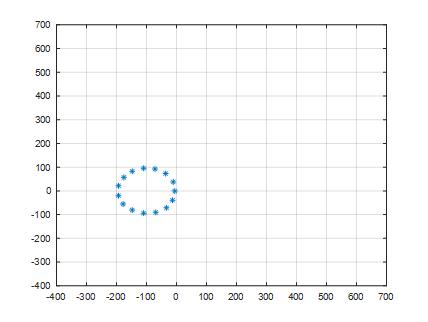}}
\caption{Sub-figures (a) shows a group of robots in a triangle formation preparing to reach a formation around a center given by the user. (b) shows the robots is moving together near to the given center. (c) shows robots are moving to reach a formation around the given center. (d) shows robots having reached a circle formation around the center. } \label{Fig:PmaxEffects}
\end{figure}

\subsubsection{Reaching a square formation from random initial positions}
In Fig. 6(a), robots' initial positions are randomly distributed. By applying methods in section \uppercase\expandafter{\romannumeral3}, the system selects a leading robot and arranges each robot in a position of the formation. Then robots move and reach the square formation. Finally, the square formation is reached as shown in Fig. 6(d).

In this example, the estimated cost is 72000 and the practical cost is 150860. In fact, when reaching the square formation, each robot chooses another one in the nearby position as the reference. It adjusts to stand at the right position relative to the reference. For the robot referred to is also moving, the destination keeps moving as well. Then the robot's practical path is far from straight. That is why the practical path causes an extra distance compared to the estimated one resulting from calculating by using straight line distance.

\subsubsection{Conversion from square formation to circle formation}
The process of transformation from square formation to circle formation is shown in Fig. 7. The user doesn't give the center. The system determines the center's position according to Theorem 2. Then the system arranges each robot to a position in the circle formation. Finally, robots move and reach the circle formation as shown in Fig. 7(d).

In the example, the estimated cost is 27211, while the practical moving cost is 27470. In the process, each robot moves referring to the fixed center. Regardless of the unfrequent collision avoidance, the practical path is nearly straight. Therefore the estimated cost is a little larger than the practical one. The collision avoidance also results in an extra moving cost.

\subsubsection{Conversion from circle formation to triangle formation}
In Fig. 8, robots transform gradually from circle formation to triangle formation. There is no center given by the user. The system first determines the center of the formation and obtains the triangle formation according to section \uppercase\expandafter{\romannumeral7}.A. Then the system completes arrangement of robots in the formation. Finally, robots move and reach the triangle formation as shown in Fig. 8(d).

In this example, the estimated cost is 15642, while the practical cost is 15652. During this transformation, there is nearly no collision avoidance. Therefore the practical cost is nearly the same as the estimated one.

\subsubsection{Reaching a circle formation around a faraway center}
In Fig. 9(a), robots are in the triangle formation. User gives a center (-100,0). Robots first move together towards the center as shown in Fig. 9(b). When they are near the desired positions, the system calculates the arrangement of robots. Then robots move to reach the circle formation as shown in Fig. 9(c). Finally they reach the circle formation in Fig. 9(d).

In this example, the estimated cost is $2.71 \times 10^{6}$, while the practical moving cost is $2.76 \times 10^{6}$. This includes the extra cost caused by collision avoidance. Furthermore, robots move near the center rather than move directly towards the right position relative to the center. This causes an extra cost compared with the straight distance.

\subsection{Moving Cost comparison}
To confirm the advantage of our proposed position arrangement strategy in reducing the moving cost, we compare it with other strategies. One strategy is called fixed position moving mode, where each robot is arranged in a fixed position in the formation. Another one is referred to as random position moving mode, where robots are arranged randomly in the formation. We do simulations and obtain the moving cost for each method. The comparison is done for two different cases. We first do experiments of reaching a square formation with a leader from random initial positions. Then we simulate the case of reaching a circle formation with a center. For each case, we do 200 experiments and compare the moving cost of the three strategies. In this simulation, our system parameter is the same as the one in Section \uppercase\expandafter{\romannumeral8}.A.
\subsubsection{moving cost for square formation}
As shown in Fig. 10, the moving cost of our proposed position arrangement is lower than the other two methods. Furthermore, its moving cost is more stable. Therefore the proposed position arrangement based on the Hungarian method can improve the system performance both in moving cost and in stability.

\begin{figure}[!h]
  \centering
  \includegraphics[width=3in]{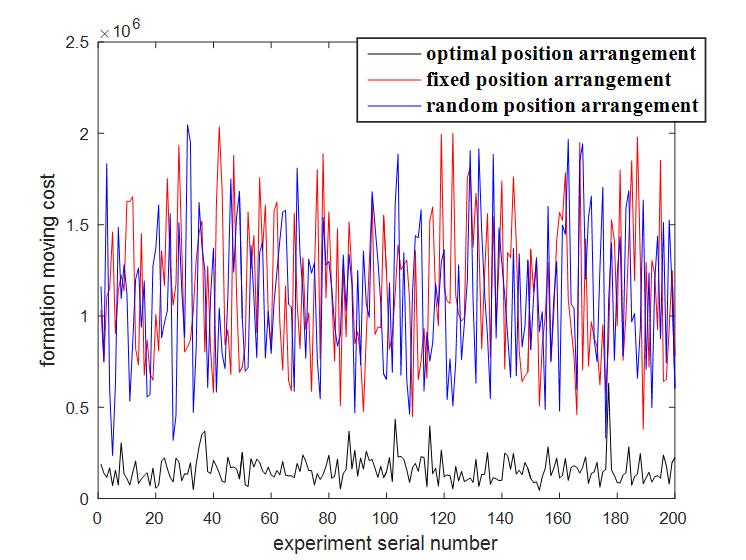}
  \caption{ Comparison of the moving cost of different position arrangement strategies for reaching the square formation. The graph represents moving cost of the three methods in different experiments. }\label{}
\end{figure}
\begin{figure}[!h]
  \centering
  \includegraphics[width=3in]{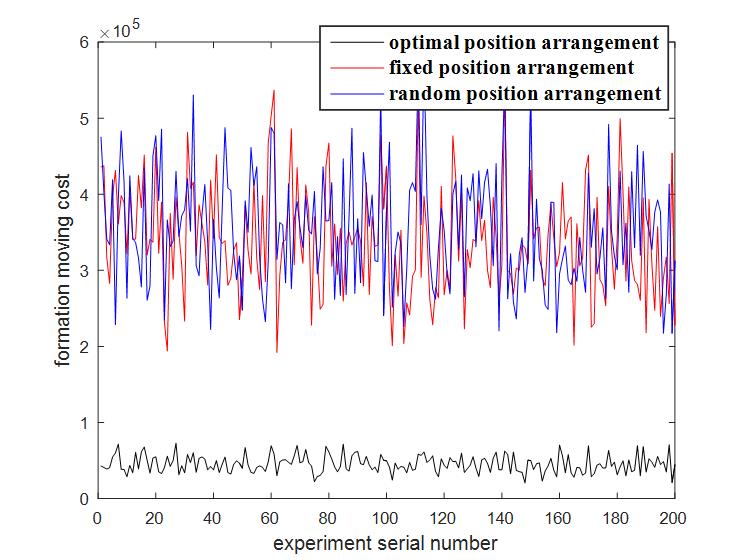}
  \caption{ Comparison of the moving cost for different position arrangement strategies for reaching a formation around a center. The graph represents moving cost of the three methods in different experiments.}\label{}
\end{figure}
\subsubsection{moving cost for circle formation}
As shown in Fig. 11, in the case of reaching the circle formation with a center, the result is similar as that presented in Fig. 10. Our arrangement strategy based on the Hungarian method can improve the system performance compared to other two methods.

\subsection{Formation bias analysis}

To investigate the offsets of the theoretical lower bound on the formation bias for the circle formation process, we carry out simulations to investigate the change of formation with respect to three different system parameters. The experimental data is fitted with the least squares method. The calculated lower bound is also depicted.

In each simulation, we adjust the conditioning parameter in a previously set range. In each parameter value, we carry out 20 experiments. We let robots be in random initial positions and then complete reaching the circle formation to obtain the formation bias for each experiment.

\begin{figure}[tbp]
\centering
\subfigure[] { \label{Fig:pm108}
\includegraphics[width=3in]{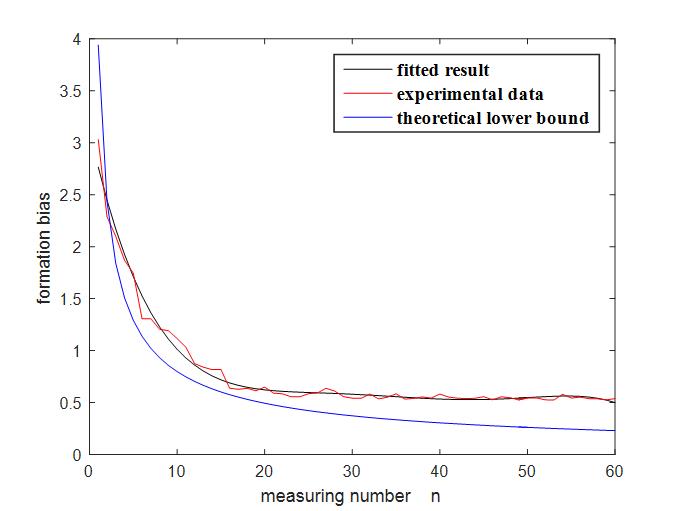}}
\subfigure[]{ \label{Fig:pm1077}
\includegraphics[width=3in]{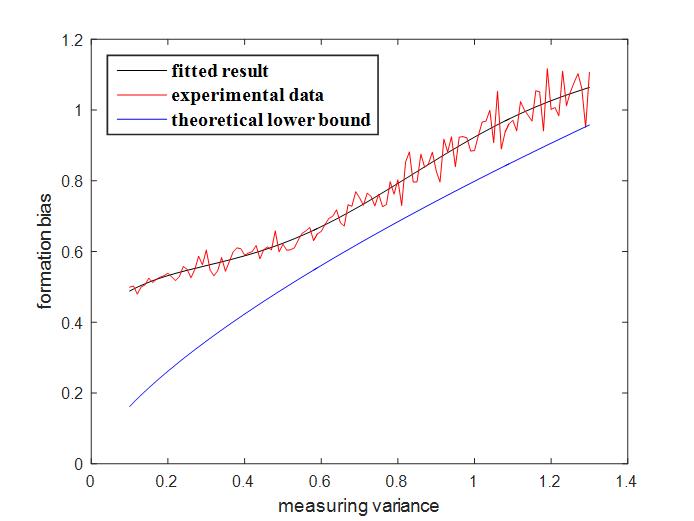}}
\subfigure[] { \label{Fig:pm1073}
\includegraphics[width=3in]{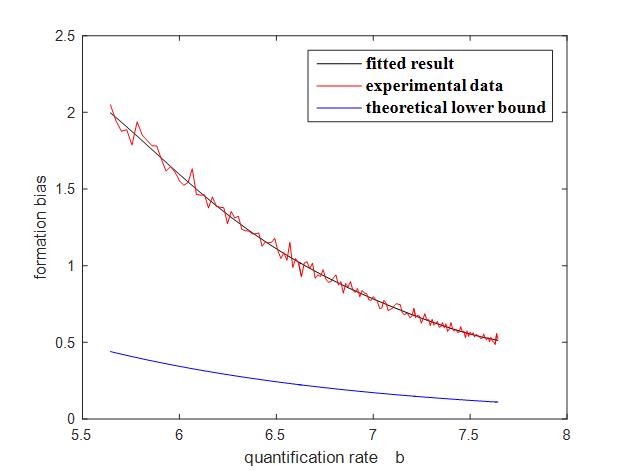}}
\caption{Sub-figures (a) shows the formation bias decreases as the measuring number $n$ increases. It depicts the experimental data, the fitted curve and the calculated lower bound. (b) shows the formation bias increases as the distance's measuring variance increases. (c) shows the formation bias decreases as the quantification rate b increases.  } \label{Fig:PmaxEffects}
\end{figure}

\subsubsection{Measuring times n}
In this simulation, the formation area is $S=28800$ and the measuring variance of angle is 0.05. There are 15 robots whose measuring variance of distance is 2. The radius of the system's controlling circle $R_{m}$ is 200. The distance's partition number is 200. Converting it to the quantification rate, we have $b=7.64$. The measuring number ranges from 1 to 60. In theoretical calculation, the distance's empirical ranging threshold is $l_{0}=120$.

Fig. 12(a) shows the change of formation bias with respect to measuring times $n$. All curves decrease as $n$ increases. From the point of common sense, when $n$ is small, increasing the measuring times can reduce the formation bias apparently. When $n$ is large enough, increasing $n$ can't obtain obvious evolution. The tendency of curves is consistent with common sense. The red curve represents experimental data and the black curve is the fitting result with least square method. The blue curve is the calculated lower bound according to Theorem 1. In the beginning, the theoretical result is a little larger than experimental data. This is due to less statistics be done. However, in the rest majority part, the lower bound is below the experimental data and the tendency is the same.

\subsubsection{Distance measuring variance $\sigma$}
In this simulation, the formation area is $S=28800$ and the measuring variance of angle is 0.05. There are 15 robots. Each robot measures the distance for 10 times. The radius of the system's controlling circle $R_{m}$ is 200. The distance's partition number is 200. Converting it to the quantification rate, we have $b=7.64$. The distance's measuring variance ranges from 0.1 to 1.3. In theoretical calculation, the distance's empirical ranging threshold is $l_{0}=120$.

Fig. 12(b) shows the change of formation bias with respect to distance measuring variance $\sigma$. In general, all curves are increasing with $\sigma$. From the point of common sense, when the measuring variance increases, the measuring bias increases and causes the formation bias to increase. The tendency of curves is consistent with common sense. The theoretical lower bound is below the experimental data and the tendency is the same.

\subsubsection{Quantification rate $b$}
In this simulation, the formation area is $S=28800$ and the measuring variance of angle is 0.05. There are 15 robots. Each robot measures the distance for 10 times. The radius of the system's controlling circle $R_{m}$ is 200. The measuring variance of distance is 0.01. The distance's partition number ranges from 50 to 200. Converting it to the quantification rate, we have $b \in [5.64,7.64]$. In theoretical calculation, the distance's empirical ranging threshold is $l_{0}=120$.

Fig. 12(c) shows the change of formation bias with respect to the quantification rate $b$. All curves increase with $b$. From the point of common sense, when $b$ increases, the measuring bias decreases, causing the formation bias to decrease. When $b$ is large enough, the decreasing tendency slows down. The tendency of curves is consistent with common sense. The lower bound is below the experimental data. Though there is a gap, the tendency is the same.

\section{Conclusion}
In this paper, we discussed the control of multi-robot formation and analyzed the system theoretically. First, we presented a method to arrange robots to positions in a formation by optimizing the total moving distance. We formulated the problem into a task distribution optimization problem and solved it in Hungarian method. Then we employed a leader-follower method to achieve the formation. Afterwards, we defined the formation bias and presented a lower bound of it, which can help us determine the parameter of the system. Besides, we also discussed control in formation changing. We also carried out various simulations to investigate the effects of system parameters on the formation bias.


%

%



\ifCLASSOPTIONcaptionsoff
  \newpage
\fi



\bibliography{usart}
\bibliographystyle{unsrt}

\end{document}